\documentclass{ieeeaccess}
\usepackage{cite}
\usepackage{amsmath,amssymb,amsfonts}
\usepackage{algorithmic}
\usepackage{graphicx}
\usepackage{textcomp}
\usepackage{url}

\def\BibTeX{{\rm B\kern-.05em{\sc i\kern-.025em b}\kern-.08em
    T\kern-.1667em\lower.7ex\hbox{E}\kern-.125emX}}
\begin{document}
\history{Received 14 July 2023, accepted 1 August 2023, date of publication 7 August 2023, date of current version 17 August 2023.}
\doi{10.1109/ACCESS.2023.3302892}

\title{Data vs. Physics: The Apparent Pareto Front of Physics-informed Neural Networks}
\author{
\uppercase{Franz M.~Rohrhofer}\authorrefmark{1}, 
\uppercase{Stefan Posch}\authorrefmark{2},
\uppercase{Clemens Gößnitzer}\authorrefmark{2}, and 
\uppercase{Bernhard C.~Geiger}\authorrefmark{1},
\IEEEmembership{Member, IEEE}}

\address[1]{Know-Center GmbH, Research Center for Data-Driven Business \& Big Data Analytics, Sandgasse 36/4, 8010 Graz, Austria}
\address[2]{LEC GmbH, Large Engines Competence Center, Inffeldgasse 19, 8010 Graz, Austria}

\tfootnote{This work was supported by the the Austrian COMET — Competence Centers for Excellent Technologies — Programme of the Austrian Federal Ministry for Climate Action, Environment, Energy, Mobility, Innovation and Technology, the Austrian Federal Ministry for Digital and Economic Affairs, and the States of Styria, Upper Austria, Tyrol, and Vienna for the COMET Centers Know-Center and LEC EvoLET, respectively. The COMET Programme is managed by the Austrian Research Promotion Agency (FFG)}

\markboth
{Franz M.~Rohrhofer \headeretal: Data vs. Physics: The Apparent Pareto Front of PINNs}
{Franz M.~Rohrhofer \headeretal: Data vs. Physics: The Apparent Pareto Front of PINNs}

\corresp{Corresponding author: Franz M.~Rohrhofer (e-mail: frohrhofer@acm.org).}

\begin{abstract}
Physics-informed neural networks (PINNs) have emerged as a promising deep learning method, capable of solving forward and inverse problems governed by differential equations.
Despite their recent advance, it is widely acknowledged that PINNs are difficult to train and often require a careful tuning of loss weights when data and physics loss functions are combined by scalarization of a multi-objective (MO) problem.
In this paper, we aim to understand how parameters of the physical system, such as characteristic length and time scales, the computational domain, and coefficients of differential equations affect MO optimization and the optimal choice of loss weights.
Through a theoretical examination of where these system parameters appear in PINN training, we find that they effectively and individually scale the loss residuals, causing imbalances in MO optimization with certain choices of system parameters.
The immediate effects of this are reflected in the apparent Pareto front, which we define as the set of loss values achievable with gradient-based training and visualize accordingly.
We empirically verify that loss weights can be used successfully to compensate for the scaling of system parameters, and enable the selection of an optimal solution on the apparent Pareto front that aligns well with the physically valid solution.
We further demonstrate that by altering the system parameterization, the apparent Pareto front can shift and exhibit locally convex parts, resulting in a wider range of loss weights for which gradient-based training becomes successful. 
This work explains the effects of system parameters on MO optimization in PINNs, and highlights the utility of proposed loss weighting schemes.
\end{abstract}


\titlepgskip=-21pt

\maketitle

\section{Introduction}\label{sec:introduction}
Recent developments in scientific computing have led to deep learning approaches that can model the dynamics of physical systems governed by differential equations.
State-of-the-art methods often infer the dynamics during model training by leveraging data that embodies the fundamental laws of physics~\cite{udrescu2020},\cite{brunton2016},\cite{sanchez2020}.
In contrast, physics-informed neural networks (PINNs) directly encode the governing differential equations as soft constraints via a physics loss function~\cite{lagaris1998}\cite{raissi2019}.
PINNs enable a seamless integration of data and physics with their respective losses often considered as multi-objective (MO).
The large-scale flexibility of PINNs together with their time-continuous, mesh-independent, and unsupervised encoding of differential equations has propelled PINNs into a vast number of multi-scale and multi-physics applications~\cite{raissi2020}\cite{chen2020}.
Today, PINNs are widespread in diverse scientific and engineering disciplines, such as bioengineering~\cite{sahli2020}\cite{kissas2020}, aerodynamics~\cite{mao2020}\cite{dourado2020}, and materials science~\cite{he2020}\cite{yin2021}.

Setting up a well-working PINN application, however, is not straightforward.
In particular, the vanilla implementation of PINNs is known to be prone to training failures that often lead to inaccurate and nonphysical predictions~\cite{krishnapriyan2021}.
The discussion on training failures in PINNs is diverse, and each problem setup seems to present its own unique challenges for PINN optimization~\cite{monaco2023}. 
This diversity makes it difficult to choose the right remedy in the face of certain optimization issues.
In general, any improvement in the robustness and generalizability of PINNs requires an understanding of the optimization complexity of the physics loss function.
As a general rule, its complexity increases as the physical system and governing differential equations become more complex. 
This is true of systems with highly-nonlinear~\cite{fuks2020}, chaotic~\cite{steger}, or multiscale dynamics~\cite{karniadakis2021}.
To cope with complex systems and geometries, domain decomposition or sequence-to-sequence methods have been developed that divide the original problem into smaller subdomains~\cite{fuks2020}\cite{jagtap2021}. 
Each subdomain is then tackled by a separate PINN, resulting in an overall lower optimization complexity.
Furthermore, soft attention mechanisms were introduced to focus the physics loss optimization on regions which are typically hard to resolve, such as discontinuities~\cite{wu2023} and stiff dynamics~\cite{mcclenny2023}. 
For systems that are slightly more complex than a previously solved problem, curriculum regularization provides a simple starting point for the PINN optimization, which gradually becomes more complex as the PINN is trained~\cite{fuks2020}.
Yet even with simple systems, the optimization may converge to suboptimal solutions that describe trivial solutions to the physics loss function~\cite{rohrhofer2023}.
In this regard, learning in sinusoidal space~\cite{wong2022} or methods that respect causality~\cite{wang2022} provide a potential remedy. 

While all the above mentioned circumstances and proposed modifications particularly apply to optimization issues related to the physics loss function, a major class of discussions addresses issues related to the MO optimization in PINNs.
Loss weighting schemes are arguably the most frequently used modifications to the vanilla PINN framework.

\subsection{Loss Weighting Schemes}\label{sec:related_work}
Loss weighting schemes originate from issues related to MO optimization~\cite{wang2021}.
In PINNs, MO optimization is inherently specified by multiple loss functions which are related to either data or physics. 
In this particular context, we use the term ``data'' to specify the use of loss functions with labeled data.
Labeled data is typically used in PINNs to encode initial and boundary conditions of forward problems or to impose additional data constraints, e.g., with data coming from experiments.
With the term ``physics'', we refer to the use of loss functions that encode the governing differential equations.
These types of loss functions do not use any labeled data, as further discussed in Section~\ref{sec:PINNs}. 
Although the total number of loss functions can be reduced, e.g., by using hard constraints~\cite{peng2020}\cite{lu2021}, most PINN applications involve the use of multiple loss functions.
The standard approach to minimizing them is gradient-based optimization of a linear scalarized MO problem, given by
\begin{equation}\label{eq:MO_scalarization}
    \min_\theta \sum_{i}\alpha_i\mathcal{L}_i(\theta),
\end{equation}
with $\theta$ denoting the network weights being optimized, $\mathcal{L}_i$ the constituent loss functions and $\alpha_i$ their respective loss weights.
The vanilla PINN framework uses an unweighted scalarization, hence $\alpha=1$ for any given loss function~\cite{raissi2019}.
However, it has been frequently reported that this simple linear combination often leads to optimization failures, which are characterized by a stalled minimization of one loss and high prediction errors.
Consequently, it has become a standard procedure in PINNs to adjust loss weights, which are trimmed (often in a trial-and-error procedure) until a sufficiently accurate prediction is obtained.
Several adaptive loss weighting schemes have been introduced to overcome the cumbersome tuning of manual loss weights.
These schemes focus on what is observed during the PINN training and rely on mean gradient statistics~\cite{wang2021}\cite{jin2021}, inverse average gradient magnitudes~\cite{maddu2022}, or maximum likelihood estimations~\cite{xiang2022}.
Recently, formulations of a constrained optimization problem that use Lagrangian multipliers have also appeared in the literature~\cite{son2022}.

\subsection{Contribution}
The use of loss weighting schemes in PINNs has become an essential approach to solving convergence issues related to MO optimization and/or refining the accuracy of final network predictions.
Yet when and why the need for loss weighting schemes arises in the first place has rarely been discussed in terms of the investigated physical system and its properties. 
In this work, we therefore focus on how system parameters, such as characteristic length and time scales, the size of the computational domain, and coefficients of differential equations, affect MO optimization in PINNs.
We observe that the choice of system parameters influences the absolute scale of both data and physics residuals. 
For certain choices of system parameters, this can result in imbalanced scales of loss residuals.
This has an immediate effect on MO optimization and the ``apparent'' Pareto front, which we define as the set of loss values that are achievable with gradient-based training.
By analyzing the apparent Pareto front and final PINN predictions, we find that loss weights can be used to compensate for the scaling effects of system parameters, with their optimal choice following the prescribed trend of which loss residuals dominate the MO optimization.
Furthermore, we find that the apparent Pareto front of more balanced residuals can exhibit locally convex parts, which results in a wider range of loss weights with which gradient-based training becomes successful.
Our results provide valuable insights for MO optimization in PINNs and contribute to a better understanding of the influence of system parameters.

The remainder of this paper is arranged as follows:
In Section~\ref{sec:PINNs}, we discuss the fundamental working method of vanilla PINNs and provide the details of MO optimization and the apparent Pareto front.
In Section~\ref{sec:experimental_setup}, we introduce two physical systems on which we perform our analysis.
The two systems employ the diffusion equation and Navier-Stokes equations, which are well-known differential equations frequently used in the PINN literature.
In Section~\ref{sec:feature_scaling}, we theoretically analyze the use of feature scaling to demonstrate how and where system parameters appear in the PINN training and affect the scale of loss residuals.
In Section~\ref{sec:pareto_front}, we empirically study the apparent Pareto front and trend of optimal loss weights while altering the system parameterization.
Section~\ref{sec:discussion} provides a discussion of our findings and Section~\ref{sec:conclusion} a conclusion.

\section{Physics-Informed Neural Networks}\label{sec:PINNs}
To briefly demonstrate the fundamental working method of PINNs, we take the originally proposed PINN framework~\cite{raissi2019} and consider a parameterized partial differential equation (PDE) of the general form
\begin{equation}
    \frac{\partial}{\partial t}u(t,x) + \mathcal{F}[u; \lambda] = 0,\quad x\in\Omega, t\in[0, T],
    \label{eq:PDE_general}
\end{equation}
where $u(t,x)$ is the solution function, $\mathcal{F}[\cdot\,;\lambda]$ represents an arbitrary potentially nonlinear differential operator with parameterization $\lambda$, and $\Omega$ represents the spatial computational domain which is a subset of $\mathbb{R}^D$.
We proceed by approximating the solution function $u(t,x)$ with a fully-connected neural network $u_\theta(t,x)$ with weights $\theta$.

\subsection{Data and Physics Loss Functions}
This paper only considers well-posed problems, i.e., problems with a unique solution function $u$, which is attained by imposing sufficient initial and boundary conditions (IC and BC).
The IC and BC define the solution function at the boundary of the computational domain, i.e., at $t=0$ and $x\in\partial\Omega$, respectively, and provide a basically infinite set of labeled training data that can be sampled once prior to network training or anew at each training epoch.
The premise for encoding the IC and BC, this dataset is used by PINNs in the data loss function
\begin{subequations}\label{eq:loss_data}
\begin{align}
\mathcal{L}_\mathcal{D}(\theta)&=\frac{1}{N_\mathcal{D}}\sum_{i=1}^{N_\mathcal{D}}\left|e_i\right|^2, \mathrm{with}\label{eq:loss_function_data} \\ 
 e_i &= u_\theta(t_i,x_i)-u_i,\label{eq:residuals_data}
\end{align}
\end{subequations}
where $\mathcal{L}_\mathcal{D}$ denotes the mean squared error loss (MSE) function given by the data residuals $e_i$ which are determined from the labeled dataset $\{(t_i, x_i),u_i\}_{i=1}^{N_\mathcal{D}}$.
Details of the exact definition of IC and BC in our experiments are given in Section~\ref{sec:experimental_setup}. 

To encode the governing differential equation, PINNs make use of automatic differentiation~\cite{baydin2018} that retrieves (partial) derivatives of the neural network function at specified coordinates, commonly called collocation points.
Derivatives can be evaluated up to the order to which the neural network activation function is differentiable.
Activation functions commonly used in PINNs are thus the hyperbolic tangent (tanh), the sinusoidal (sin), or the swish.
With the neural network derivatives evaluated at the collocation points, the physics loss function is given by
\begin{subequations}\label{eq:loss_physics}
\begin{align}
\mathcal{L}_\mathcal{F}(\theta) &= \frac{1}{N_\mathcal{F}}\sum_{i=1}^{N_\mathcal{F}}\left|f_i\right|^2,\mathrm{with}\label{eq:loss_function_physics} \\ 
f_i&= \frac{\partial}{\partial t}u_\theta(t_i, x_i) + \mathcal{F}[u_\theta(t_i, x_i); \lambda],\label{eq:residuals_physics}
\end{align}
\end{subequations}
where $\mathcal{L}_\mathcal{F}$ denotes the MSE loss function given by the physics residuals $f_i$ which are determined from the unlabeled dataset $\{(t_i, x_i)\}_{i=1}^{N_\mathcal{F}}$.
In comparison to conventional PDE solvers that use a predefined computational mesh, the choice of collocation points in PINNs is not restricted, i.e., they can be randomly sampled from inside the function domain once or at each epoch, e.g., by using hyper-cube sampling.

\subsection{The Apparent Pareto Front}
Vanilla PINNs and many variants thereof handle data and physics loss functions in a MO manner.
The standard approach in PINNs is to simultaneously minimize both by defining a single or total loss function via linear scalarization of the MO problem previously introduced by~\eqref{eq:MO_scalarization}. 
The particular use of loss weights and their optimal choice differs in the literature (see Section~\ref{sec:related_work}):
they depend on the total number of loss functions in use and are tuned manually or in an adaptive weighting scheme.

From a theoretical perspective, any choice of loss weight in a scalarized problem selects a Pareto optimal solution to the MO optimization problem~\cite{hwang2012}.
The Pareto front can then be seen as the set of all possible Pareto optima that are attained by continuous adjustment of the loss weights.
In fact, any use of loss weights in a well-posed problem would converge to the same Pareto optima as the true solution of~\eqref{eq:PDE_general}, since a perfectly approximated solution function yields zero for all losses.    
However, finite-sized networks and highly complex loss functions are used in practice, so the gradient decent optimization eventually converges to local optima.
Hence these optima do not represent true Pareto optimal solutions but approximations that are found along the optimization path as selected by the particular choice of loss weights.
For our analysis, we thus define the ``apparent'' Pareto front as the set of loss values that are achievable with gradient-based optimization.
To empirically obtain the apparent Pareto front, we train several PINN instances with different loss weights and observe where the gradient-based optimization has converged (see Section~\ref{sec:pareto_front} for further details).

For the sake of simplicity and demonstration purposes, this paper only distinguishes between data and physics loss functions and introduces a single parameter that is manually tuned and simultaneously trades both by
\begin{equation}
\mathcal{L}(\theta;\alpha):=\alpha \mathcal{L}_\mathcal{D}(\theta)+(1-\alpha) \mathcal{L}_\mathcal{F}(\theta),
\label{eq:loss_weighted}
\end{equation}
where $\alpha\in(0,1)$. 
According to this definition, $\alpha \to 1$ favors low data losses, while $\alpha \to 0$ favors low physics losses.
A loss weight of $\alpha=0.5$ relates to unweighted MO optimization as used by vanilla PINNs.

\section{Experimental Setup}\label{sec:experimental_setup}

\begin{figure}[t]
\begin{center}
\centerline{\includegraphics[width=\columnwidth]{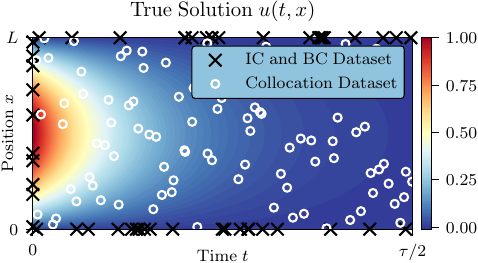}}
\caption{Geometrical setup and reference solution for the diffusion example. Representative sample of training data is shown as black crosses (IC and BC) and white circles (collocation).}    
\label{fig:system_diffusion}
\end{center}
\end{figure}

\subsection{Diffusion Example}\label{sec:problem_setup_heat}
The diffusion equation, along with variants thereof, is among the most widely studied parabolic PDEs with applications in many fields of science, pure mathematics, and engineering.
In this work we study the heat equation, a special case of the diffusion equation in the context of engineering, specifically considering the cooling of a one-dimensional rod with an initial temperature distribution and Dirichlet boundary conditions on both rod ends.
The dynamics of the system are described by the equation
\begin{equation}
    \frac{\partial }{\partial t}u(t,x)=\kappa \frac{\partial^2 }{\partial x^2}u(t,x),
    \label{eq:PDE_heat}
\end{equation}
where the solution function $u$ represents the temperature of the rod at position $x$ and time $t$, and $\kappa$ is the thermal diffusivity.
We define the initial (IC) and boundary (BC) conditions as
\begin{align}
\mathrm{IC}&: &  &u(0,x)=\text{sin}\left(\pi\frac{x}{L}\right) &  &x\in[0,L], \label{eq:IC_heat}\\ 
\mathrm{BC}&: &  &u(t,0)=u(t,L)=0   &  &t\in[0,T],  \label{eq:BC_heat}
\end{align}
with $L$ denoting the length of the rod and $T$ the simulation time.
For our later analysis and the sake of simplicity, we introduce the characteristic (diffusive) time scale 
\begin{equation}
    \tau:=\frac{L^2}{\kappa},
\end{equation}
and consider the simulation time as multiplies of it, i.e., $T=\lambda\tau$ with $\lambda\in\mathbb{R}^+_*$.

The problem stated above is well-posed and has a unique solution
\begin{equation}
u(x,t) = \text{sin}\left(\pi\frac{x}{L}\right) e^{-\pi^2 t/\tau}. 
\label{eq:solution_heat}
\end{equation}
Fixing $\lambda$ while adjusting $L$ and $\kappa$ leaves the intrinsic diffusion process and solution function unaffected.
This gives us a controllable setting of different system parameterizations all described by the same solution function (cf. Fig.~\ref{fig:system_diffusion}).
In an engineering context, this is known as dynamic similitude.

\subsection{Navier-Stokes Example}\label{sec:problem_setup_NS}
The Navier-Stokes equations are the governing PDEs in the study of fluid flow.
Along with the continuity equation, they describe the conservation of momentum and mass of viscous fluids.
Their importance in scientific and engineering modeling is indisputable as they are applied to many physical phenomena occurring in weather forecasting, blood flow through vessels, or flow around obstacles.
For our study, we limit their application to a two-dimensional incompressible steady-state flow where the solution function to be approximated by the PINN is given by the vector-valued function $(x,y)\mapsto(u, v,p)$.
The governing equations for the conservation of momentum and mass, respectively, are
\begin{subequations}\label{eq:PDE_NS}
\begin{align}
(\mathbf{u}\cdot \nabla)\mathbf{u}&=-\frac{1}{\rho}\nabla p + \nu \nabla^2\mathbf{u} \label{eq:PDE_NS_momentum}\\ 
\nabla \cdot \mathbf{u} &= 0, \label{eq:PDE_NS_mass}
\end{align}
\end{subequations}
where $\mathbf{u}=(u,v)$ and $p$ denote fluid velocity in the $x$- and $y$-directions and pressure, respectively, and $\rho$ and $\nu$ are the fluid density and viscosity.
The continuity equation~\eqref{eq:PDE_NS_mass} can be hard coded with PINNs (see~\cite{raissi2019}) which leaves behind the conservation of momemtum~\eqref{eq:PDE_NS_momentum} to be 
encoded in the physics loss function.

Analytical solutions in fluid dynamics are rare.
Therefore, we consider the laminar fluid flow in the wake of a two-dimensional grid which has been solved analytically and is better known as Kovasznay flow~\cite{kovasznay1948}.
With $L$ as the characteristic spacing of the grid, $u_0$ the mean velocity in the $\mathrm{x}$-direction and assuming constant density $\rho$, the analytical solution to this problem is given by
\begin{subequations}\label{eq:solution_NS}
\begin{align}
u(x,y) &=u_0\left(1-e^{\gamma\frac{x}{L}}\text{cos}\left(2\pi\frac{y}{L}\right)\right),\label{eq:solution_NS_u}\\
v(x,y) &=\frac{u_0\gamma}{2\pi}e^{\gamma\frac{x}{L}}\text{sin}\left(2\pi\frac{y}{L}\right),\label{eq:solution_NS_v}\\
p(x,y) &=u_0^2 e^{2\gamma\frac{x}{L}}+C\label{eq:solution_NS_p},
\end{align}
\end{subequations}
where $C$ is a constant and
\begin{equation}
    \gamma = \frac{1}{2\nu}-\sqrt{\frac{1}{4\nu^2}+4\pi^2}.
\end{equation}
To study this problem with different system parameterizations, we again make use of the concept of dynamic similitudes and consider a fixed Reynolds number $\mathrm{Re}=Lu_0/\nu$ for the above mentioned system.
Under these conditions, we adjust the mean velocity by $u_0=1/L$ (neglecting physical units) so that any selected grid spacing $L$ represents the same fluid flow.  
To establish geometric similarity, the spatial extension is assumed to be equal in both directions, i.e., $x\in[0,L]$ and $y\in[0,L]$.

In this example, the PINN is trained on boundary data sampled from the analytical solution of the flow velocities~\eqref{eq:solution_NS_u} and \eqref{eq:solution_NS_v} at the boundary.
Since the flow is steady state, there is no IC.
The pressure is entirely learned by the PINN and inferred from the set of governing PDEs~\eqref{eq:PDE_NS_momentum}.

\begin{figure}[t]
\begin{center}
\centerline{\includegraphics[width=\columnwidth]{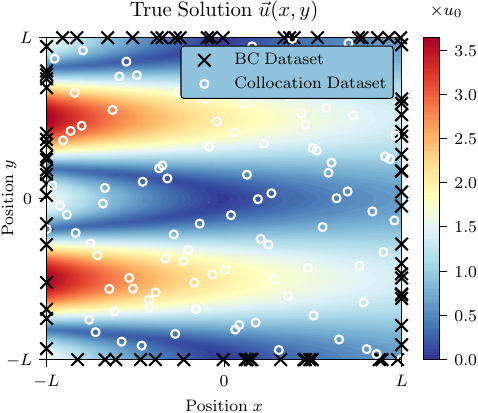}}
\caption{Geometrical setup and reference solution for the Navier-Stokes example. Representative sample of training data is shown as black crosses (IC and BC) and white circles (collocation).}    
\label{fig:system_navier_stokes}
\end{center}
\end{figure}
\section{Effects of System Parameters on Loss Residuals}\label{sec:feature_scaling}
In this section, we show that the absolute scale of data and physics residuals depends on the underlying system parameters.
We discuss the effects of feature scaling, which is a necessary step to bring the input dimensions into a range suitable for gradient descent optimization.
Our focus is on min-max feature scaling since its input ranges are well-defined and given by the computational domain. 

\subsection{Diffusion Example}
For the diffusion equation, we consider the input ranges $(t,x)\in[0,T]\times[0,L]$ scaled to the unit interval $[0, 1]^2$
\begin{equation}\label{eq:feature_scaling_heat}
    \hat t=\frac{t}{T}, \quad\mathrm{and}\quad \hat x=\frac{x}{L},
\end{equation}
where $\hat t$ and $\hat x$ denote the scaled input variables.
It is noteworthy that applying feature scaling can be seen as scaling the network function to the physical system, thus adapting to characteristic time and length scales as apparent by~\eqref{eq:feature_scaling_heat}.
The physics residuals for the diffusion equation can thus be written in terms of the scaled and characteristic quantities:
\begin{equation}
    f_i=\frac{1}{T}\frac{\partial}{\partial \hat t}u_\theta(\hat t_i, \hat x_i)-\frac{\kappa}{L^2}\frac{\partial^2}{\partial \hat x^2}u_\theta(\hat t_i,\hat x_i),
\end{equation}
where the additional pre-factors in the equation appear when the chain rule of derivatives is applied.
We again consider the simulation time as multiples of the characteristic diffusive time scale, i.e., $T=\lambda\tau=\lambda L^2/\kappa$, and rearrange the physics residuals according to this definition to yield the following:
\begin{equation}
    f_i=\frac{\kappa}{L^2} \left( \frac{1}{\lambda}\frac{\partial}{\partial \hat t}u_\theta(\hat t_i, \hat x_i)-\frac{\partial^2}{\partial \hat x^2}u_\theta(\hat t_i,\hat x_i) \right).
\end{equation}
It is now apparent that applying feature scaling re-parameterizes the diffusion equation encoded in the physics loss function and effectively scales the physics residuals by $f_i\sim \kappa/L^2$.
In contrast, the data residuals are not affected by those system parameters and stay in the range of $e_i\sim1$ due to the chosen IC~\eqref{eq:IC_heat}.
Consequently, data and physics residuals scale differently according to the scale ratio given by
\begin{equation}\label{eq:scaling_diffusion}
    f_i/e_i \sim \frac{\kappa}{L^2}.
\end{equation}
This shows how a certain choice of $L$ and $\kappa$ overbalances their scale.
While for $\kappa/L^2 \gg 1$ the scale of physics residuals is predominant, for $\kappa/L^2 \ll 1$ the scale of data residuals is predominant. 
This scaling in turn directly affects the loss values in~\eqref{eq:loss_weighted} and respective gradients with measurable consequences for MO optimization~\cite{wang2021}.

\subsection{Navier-Stokes Example}
For the Navier-Stokes system, we consider the input ranges $(x,y)\in[-L,L]^2$ scaled to the interval $[-1, 1]^2$
\begin{equation}
    \hat x=\frac{x}{L}, \quad\quad \hat y=\frac{y}{L}, \quad\mathrm{thus}\quad \hat \nabla=L\nabla.
\end{equation}
This time, the solution function will be affected by the particular choice of $L$, since $u_0=1/L$ (see experimental setup~\ref{sec:problem_setup_NS}).
Hence, we consider the velocities and pressure in terms of scaled quantities that are commonly used in the nondimensionalization of the Navier-Stokes equations:
\begin{equation}
 \mathbf{\hat u}:=\frac{\mathbf{u}}{u_0}, \quad\mathrm{and}\quad \hat p:=\frac{p}{\rho u_0^2}.
\end{equation}
Rewriting the physics residuals in terms of the scaled quantities yields
\begin{equation}
    f_i =\frac{u_0^2}{L}\left(\mathbf{\hat u}_\theta \cdot \hat\nabla\right)\mathbf{\hat u}_\theta+\frac{u_0^2}{L}\hat\nabla \hat p_\theta - \nu \frac{u_0}{L^2}\hat\nabla^2\mathbf{\hat u}_\theta,
\end{equation}
where for clarity we have omitted the spatial dependencies $\mathbf{\hat u}_\theta(\hat x, \hat y)$ and $\hat p_\theta(\hat x, \hat y)$.
After $u_0=1/L$ is used and rearranged,
\begin{equation}
    f_i =\frac{1}{L^3} \left( (\mathbf{\hat u}_\theta \cdot \hat\nabla)\mathbf{\hat u}_\theta+\hat\nabla \hat p_\theta - \nu \hat\nabla^2\mathbf{\hat u}_\theta \right)
\end{equation}
it is again apparent that the physics residuals are scaled by the factor $f_i\sim 1/L^3$, while the data residuals are scaled by $e_i\sim 1/L$ since $e_i=u_0(\mathbf{\hat{u}}_\theta(\hat{x}_i, \hat{y}_i) - \mathbf{\hat{u}}_i)$.
Consequently, data and physics residuals are scaled differently with their scale ratio given by
\begin{equation}\label{eq:scaling_navier_stokes}
    f_i/e_i \sim \frac{1}{L^2}.
\end{equation}
Again, this demonstrates that a certain choice of $L$ overbalances their scale with consequences for MO optimization.

\begin{figure*}[t]
\begin{center}
\centerline{\includegraphics[width=\textwidth]{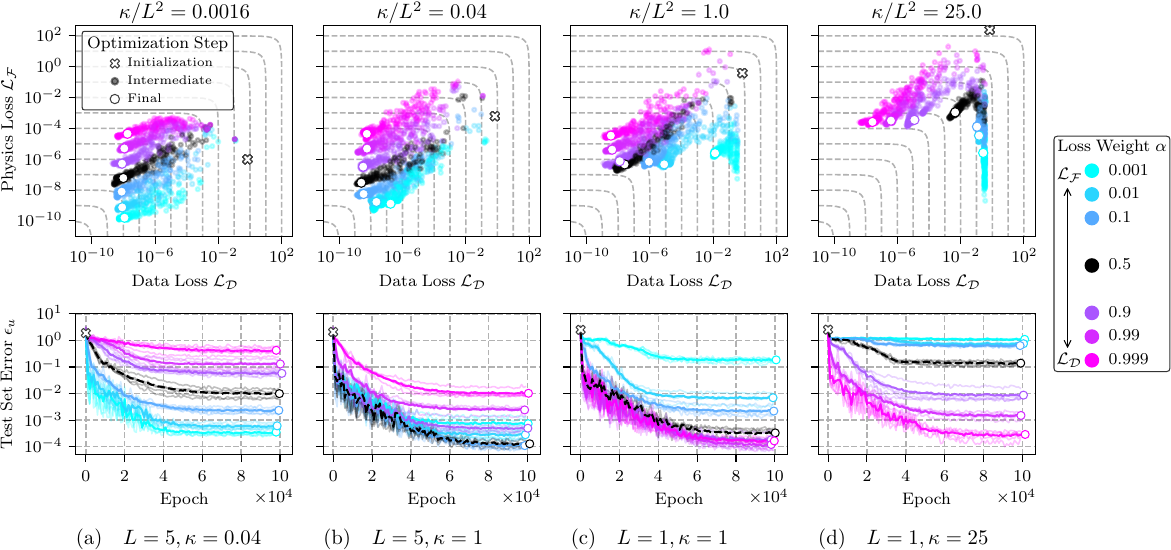}}
\caption{Data vs. physics loss (top row) and test set errors (bottom row) for the diffusion equation. Different system parameterizations are arranged as columns. The system parameters determine the residual scale ratio $f_i/e_i\sim\kappa/L^2$, where for $\kappa/L^2 \gg 1$ the scale of physics residuals is predominant and for $\kappa/L^2 \ll 1$ that of data residuals. Final optimization steps (empty circles) in the top row outline the apparent Pareto front.}    
\label{fig:pareto_diffusion}
\end{center}
\end{figure*}
\section{Studying the Apparent Pareto Front and Optimal Choice of Loss Weights}\label{sec:pareto_front}
In this section, we empirically analyze the scaling effects of system parameters on MO optimization of data and physics.
To understand the qualitative trend of optimal loss weights, we analyze the apparent Pareto front with different system parameterizations and study which loss weights achieve a good PINN performance in terms of the relative $\ell^2$ error.  
We use the concept of dynamic similitudes (see experimental setup~\ref{sec:experimental_setup}) to study one and the same physical system with individual sets of system parameters, which causes different scaling effects captured by the residual scale ratio $f_i/e_i$.

\subsection{PINN Setup}\label{sec:problem_setup_PINN}
We optimize PINNs with four hidden layers and 50 neurons per layer, employing a hyperbolic tangent (tanh) activation function for the hidden layers and a linear activation for the output layer\footnote{All code is available on GitHub at~\url{https://github.com/frohrhofer/PINN_pareto}}.
Network weights are initialized using the Glorot uniform~\cite{glorot2010} initializer.
We use Adam~\cite{kingma2014} to minimize the MO loss~\eqref{eq:loss_weighted} with a learning rate of $0.01$.
Training is performed for $10^5$ epochs to ensure that the optimization has converged to a solution, where measured quantities have stabilized and no longer change substantially. 
Training data is sampled anew at each epoch with $N_\mathcal{F}=1024$ (unlabeled collocation points) and $N=128$ (labeled data points) at each boundary, thus $N_\mathcal{D}=384$ for the diffusion example (cf. Fig.~\ref{fig:system_diffusion}) and $N_\mathcal{D}=512$ for the Navier-Stokes example (cf. Fig.~\ref{fig:system_navier_stokes}).

To evaluate the accuracy of predictions, we measure the relative $\ell^2$ error on an independent test set, which is randomly sampled inside the computational domain with $N=1024$.
The relative $\ell^2$-error is given by
\begin{equation}
    \epsilon_u \equiv \frac{\|u-u_{\theta}\|_2}{\|u\|_2},
\end{equation}
where $u$ is obtained from the analytical solutions~\eqref{eq:solution_heat} and~\eqref{eq:solution_NS}.
For qualitative analysis of the apparent Pareto front, we train PINN instances with loss weights $\alpha=\{0.001, 0.01, 0.1,0.5,0.9,0.99,0.999\}$.
Test runs are repeated with five unique seeds for data sampling and network initialization.

Note that we have also performed tests with different activation functions (sin, swish), network architectures (2x30, 6x100), learning rates (0.001, 0.0001), and training set sizes, but the results had a similar qualitative outcome as presented in the subsection.
For the sake of simplicity, the following subsections are concerned with only the setting presented above.

\subsection{Diffusion Example}\label{sec:results_heat}
The diffusion example is parameterized by $L$, $\kappa$, and $\lambda$, where for a fixed value of $\lambda$ any choice of $L$ and $\kappa$ represents the same underlying system dynamics as with the analytical solution given by~\eqref{eq:solution_heat}, cf. Fig.~\ref{fig:system_diffusion}. 
In this example, the residual scale ratio is determined by the system parameters through $f_i/e_i \sim \kappa/L^2$.
To study the effects on the apparent Pareto front and the optimal choice of loss weights, we thus choose $\lambda=0.5$ for our experiment and take $(L,\kappa)\in\{(5,0.04), (5,1), (1,1), (1,25)\}$ to cause a respective scaling by $\kappa/L^2\in\{0.0016,0.04,1,25\}$.
We have also performed a test with $\lambda\in\{0.1,1,10\}$, but its results revealed qualitative behavior similar to what is presented in this section.

Fig.~\ref{fig:pareto_diffusion} shows the results of this experiment; 
the different system parameterizations are arranged by column.
The top row displays the history of loss tuples ($\mathcal{L}_\mathcal{D}$ and $\mathcal{L}_\mathcal{F}$) in each of the trained PINN instances as recorded during the training. 
To increase clarity and comprehension, initial (epoch $0$) and final (epoch $10^5$) optimization steps are averaged across the five different PINN runs and highlighted as empty crosses and circles, respectively.
It is clear that each PINN instance starts close to the marked initialization step and eventually converges to different regions as determined by the loss weight $\alpha$.
The apparent Pareto front can thus be seen as a theoretical interpolation curve of the final loss values indicated by the empty circles.
Note that as a result of utilizing a stochastic optimization method, it is possible for those points to have slightly worse values than those found in intermediate optimization steps.
The bottom row of the figure shows the respective test set errors ($\epsilon_u$) as a function of the optimization step.
The test set errors are again averaged over the five PINN instances using different seeds, which are presented by thick lines while individual runs are thin.
The black dashed line highlights the unweighted MO, and the empty circles at the end of the test error curves are randomly shifted vertically to reduce overlap.

Looking at the test set error in the bottom row of the figure, we observe a common trend from left to right:
in the presence of predominant data residuals (Fig.~\ref{fig:pareto_diffusion}a), a higher weighting of the physics loss ($\alpha\to0$) yields low test set errors.
On the contrary, when predominantly physics residuals are present (Fig.~\ref{fig:pareto_diffusion}d), a higher weighting of the data loss ($\alpha\to1$) is necessary to achieve comparable low errors. 
Intermediate scale factors (Fig.~\ref{fig:pareto_diffusion}b and \ref{fig:pareto_diffusion}c) follow this trend where the unweighted optimization ($\alpha=0.5$) already achieves comparably low errors due to the more balanced residual scales.
Yet when $\kappa/L^2=1$, performance is slightly better when a higher weight is given to data loss.

The apparent Pareto front in the top row provides further insights:
in general, we observe the apparent Pareto front substantially changing its shape with different system parameterizations.
In Fig.~\ref{fig:pareto_diffusion}b and~\ref{fig:pareto_diffusion}c, we find that the apparent Pareto front in the lower left region locally exhibits  convex parts (not explicitly highlighted), where generally a wider range of loss weights achieves similarly accurate results, i.e., MO optimization is less sensitive to the particular choice of the loss weight $\alpha$.
In contrast are the apparent Pareto fronts in the presence of unbalanced residual scales (Fig.~\ref{fig:pareto_diffusion}a and~\ref{fig:pareto_diffusion}d), where only a particular choice of $\alpha$, here either $\alpha\to0$ or $\alpha\to1$, yields accurate results.

\begin{figure*}[t]
\begin{center}
\centerline{\includegraphics[width=\textwidth]{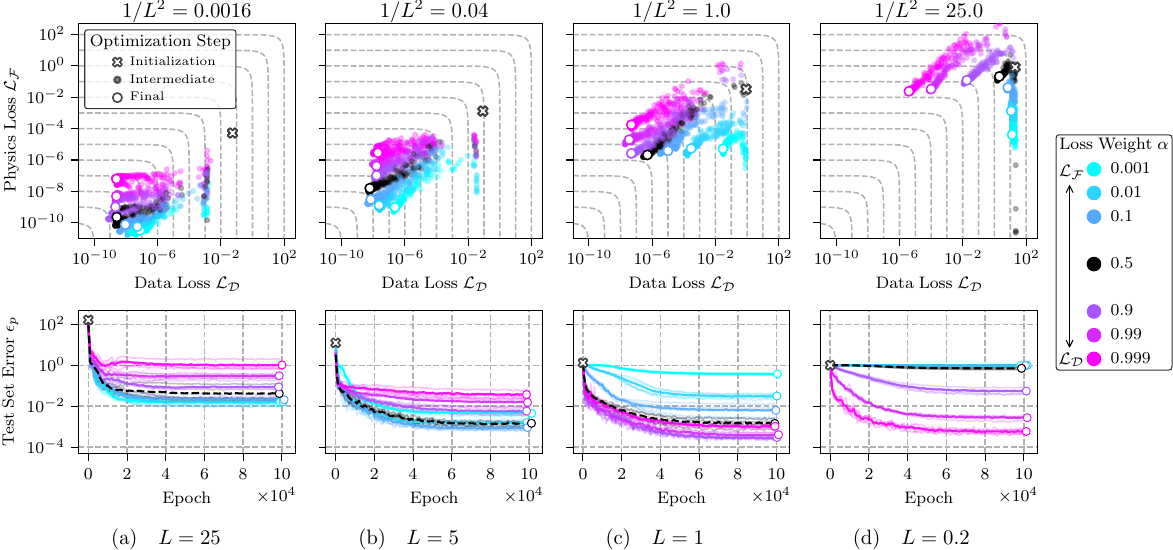}}
\caption{Data vs. physics loss (top row) and test set errors (bottom row) for the Navier-Stokes equations. Different system parameterizations are arranged as columns. The system parameters determine the residual scale ratio $f_i/e_i\sim1/L^2$, where for $1/L^2 \gg 1$ the scale of physics residuals is predominant and for $1/L^2 \ll 1$ that of data residuals. Final optimization steps (empty circles) in the top row outline the apparent Pareto front.}    
\label{fig:pareto_navier_stokes}
\end{center}
\end{figure*}
\subsection{Navier-Stokes Example}\label{sec:results_navier_stokes}
The Navier-Stokes setup has a single system parameter $L$ that simultaneously affects both residual scales;
their scale ratio is determined by $f_i/e_i\sim1/L^2$.
We thus choose $L\in\{25,5,1,0.2\}$ to cause a respective scaling by $1/L^2\in\{0.0016,0.04,1,25\}$.

The history of loss tuples ($\mathcal{L}_\mathcal{D}$ and $\mathcal{L}_\mathcal{F}$), and test set errors ($\epsilon_p$) for this experiment is found in Fig.~\ref{fig:pareto_navier_stokes}.
The same explanation of the figure that was provided in the previous subsection also applies here.
Note that since the relative $\ell^2$ error is dependent on the absolute range of the target value and the range in this example varies with regard to different system parameterizations, a direct quantitative comparison of test errors between different settings of $L$ is not possible.
Furthermore, since the velocity errors ($\epsilon_u$ and $\epsilon_v$) exhibit behavior qualitatively similar to that of the pressure field, their display is omitted for the sake of clarity.

Similar to the previous example, a common trend is observed from left to right:
in the presence of predominant data residuals (Fig.~\ref{fig:pareto_navier_stokes}a), a higher weighting of the physics loss ($\alpha\to0$) achieves lower prediction errors.
In contrast, the setting with predominantly physics residuals (Fig.~\ref{fig:pareto_navier_stokes}d) requires higher weighting of the data loss ($\alpha\to1$) to yield accurate results.
The trend is similar with intermediate settings (Fig.~\ref{fig:pareto_navier_stokes}b and \ref{fig:pareto_navier_stokes}c), where the unweighted MO already achieves accurate results when $1/L^2=0.04$, while a slightly higher weighting of the data loss works better when $1/L^2=1$.

Regarding the apparent Pareto front, we once again observe that its shape is substantially influenced by the chosen parameterization.
As observed in the previous example, locally convex parts emerge in the lower left corner of the apparent Pareto front (here in Fig.~\ref{fig:pareto_navier_stokes}a-c), and this coincides with a wider range of loss weights that achieve similar accuracy.

\section{Discussion}\label{sec:discussion}
It has become widely known that the success of PINN training in the sense of obtaining the physically correct solution depends on a myriad of factors, and hence requires a variety of mitigation methods. 
In general, these factors can be categorized as depending on model parameters and system parameters, though we believe that this separation is less clear than in classical deep learning problems.

This paper has shed more light on the effect of system parameters. 
Specifically, Section~\ref{sec:feature_scaling} shows mathematically with two systems of differential equations that changing the size of the computational domain, e.g., by altering the system parameterization, affects the loss residuals used in MO optimization. 
Our experiments in Section~\ref{sec:pareto_front} are in line with our mathematical analysis and show how varying the system parameters leads to different scaling factors of loss residuals, which in turn requires different loss weights $\alpha$ for PINN training to be successful.
Altering the system parameterization thus effectively converts one specific PINN training instance to another that is characterized by different loss weights.\footnote{Additional investigations, which are not reproduced here, have revealed that the new PINN training instance may also be characterized by differently parameterized differential equations.} 
This insight resonates with and mathematically supports the literature that claims that successful PINN training requires carefully selected loss weights or proposes automatic loss weighting schemes. 
While proposed methods are usually tailored to specific cases and directly address issues as they arise during training, our results offer a more comprehensive view and focus on potential root causes of issues that may arise from inappropriate system parameterizations.

From the perspective of differential equations, the computational domain is rarely assumed to be fixed. 
Instead and especially in computational fluid dynamics problems are often nondimensionalized or scaled. 
In other words, the parameters of the system under consideration are reduced to their minimum number, and often normalized with regard to intrinsic quantities (e.g., the spatial domain is normalized with regard to a characteristic length scale). 
Nondimensionalization therefore suggests a certain set of system parameters and subsequently a set of ideal loss weights. 
Note that the ideal loss weights still have to be discovered, even those in nondimensionalized systems: 
In Fig.~\ref{fig:pareto_navier_stokes} $(c)$, it can be seen that the best test set error is achieved with a loss weight of $\alpha=0.9$.

From a practical perspective, we believe that the insights from this work can help to improve PINN training. 
For example, suppose that a parameterization (nondimensionalized or not) and PINN configuration (network architecture, activation function, loss weights, etc.) are known that achieve satisfactory training performance with a given system of differential equations.
Such configurations may be taken from existing literature that has successfully applied PINNs to problems of practical relevance. 
To solve an instance of the same problem with different system parameters, the results presented here can now be taken to study how the individual loss terms of the new problem scale relative to the original problem as a function of the changed system parameters. 
This can then be used to adjust the loss weights of the new PINN instance. 
(Yet note that with this loss weight adjustment, the parameters of the differential equation may also change relative to the original and new systems.) 
Supposing that each differential equation and each dimension of the IC and BC losses is assigned a separate loss weight, a certain number of system parameter changes can be compensated for. 
If this does not suffice to ensure successful PINN training, one can resort to other methods that directly affect the system parameters (see Section~\ref{sec:related_work}). 
These include domain decomposition (which either explicitly or indirectly, via adaptive collocation point sampling, affects the effective size of the computational domain), and curriculum learning (which starts PINN training at ``simple'' parameterizations and slowly changes the parameters to those of the new system). 
Moreover, when dealing with solution functions that have challenging regions to learn, such as discontinuities or stiff systems, the physics residuals can become comparably larger in these regions and exceed the scaling effects of system parameters. Therefore, it may become necessary to weight individual collocation points to handle such situation effectively.
Future research should investigate whether such a general procedure makes PINN training more reliable; this paper has taken an important step in this direction.

\section{Conclusion}\label{sec:conclusion}
In this paper, we highlighted the importance of understanding the role of system parameters in the training of PINNs. 
We demonstrated that system parameters such as characteristic length and time scales, the computational domain, and coefficients of differential equations influence the absolute scale of data and physics residuals.
We theoretically and empirically verified that this has a consequence for PINN training, especially when data and physics loss functions are minimized through a scalarized MO optimization problem.
As measured by the set of loss values that are achievable with gradient-based training and defined as the ``apparent'' Pareto front, we observed that its shape and the optimal choice of loss weights are determined by the chosen system parameters. 
Moreover, we showed by altering the system parameterization that the apparent Pareto front can exhibit locally convex parts, resulting in a wider range of loss weights for which gradient-based training is successful.
These findings provide important insights into the MO optimization of PINNs and suggest that the common practice of nondimensionalization can have significant implications for PINN training. 
Ultimately, this work contributes to the development of more effective and efficient loss weighting schemes, which take into account fundamental properties of physical systems governed by differential equations.


\begin{IEEEbiography}[{\includegraphics[width=1in,height=1.25in,clip,keepaspectratio]{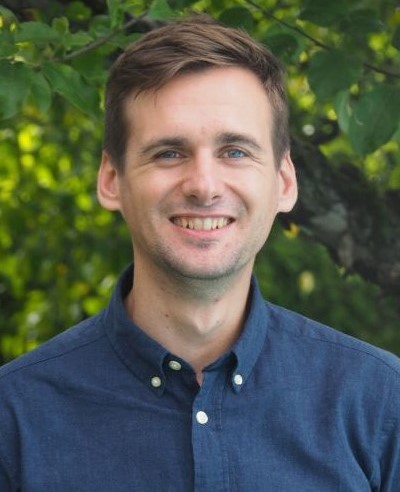}}]{Franz M. Rohrhofer} received a B.Sc. in physics from the University of Graz, Austria, in 2016 and a Dipl.-Ing. degree in technical physics (with distinction) from Graz University of Technology, Austria, in 2019.
From 2019 to 2020 he worked as a Project Assistant at the Institute of Theoretical and Computational Physics at Graz University of Technology, Austria.
In this position, his work included the development of machine learning surrogates for crystal structure prediction with a special focus on hybrid and theory-assisted feature engineering.
In 2020 he joined Know-Center GmbH, Graz, Austria, as a Research Scientist and started a PhD programme in computer science at Graz University of Technology, Austria. 
His current work and research interests comprise the development of physics-informed deep learning approaches for computational fluid mechanics and reactive flow simulations.
\end{IEEEbiography}

\begin{IEEEbiography}[{\includegraphics[width=1in,height=1.25in,clip,keepaspectratio]{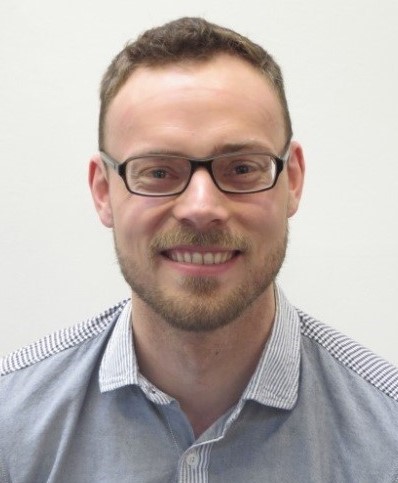}}]{Stefan Posch} received a B.Sc. (2011), M.Sc. (2013) and Ph.D. (2017) in mechanical engineering at Graz University of Technology, Austria. 
He then worked as a senior engineer at Midea Austria GmbH where he was responsible for simulation tasks in the field of hermetic compressors.
Since 2019 he has worked at the Large Engines Competence Center GmbH in Graz, Austria, as a senior scientist and team leader for system simulation and AI integration. 
His main research interests include the combination of numerical simulation and data-driven approaches
\end{IEEEbiography}

\begin{IEEEbiography}[{\includegraphics[width=1in,height=1.25in,clip,keepaspectratio]{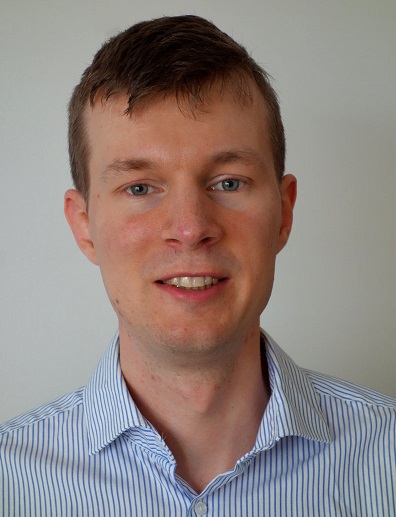}}]{Clemens Gößnitzer} received his bachelor’s (2014), master’s (2016), and doctoral (2019) degrees in chemical and process engineering from Vienna University of Technology, Austria. 
After his Ph.D., he joined the Large Engines Competence Center GmbH in Graz, Austria, as a senior engineer, where he works on ignition modeling and combustion simulation in the context of internal combustion engines.
Since 2021 he has led the CFD team and is responsible for the physical modeling and simulation of complex combustion phenomena. 
His research interests are the simulation of reactive flows with conventional and emerging fuels, physical modeling, and the integration of machine learning approaches in physical simulations.
\end{IEEEbiography}

\begin{IEEEbiography}[{\includegraphics[width=1in,height=1.25in,clip,keepaspectratio]{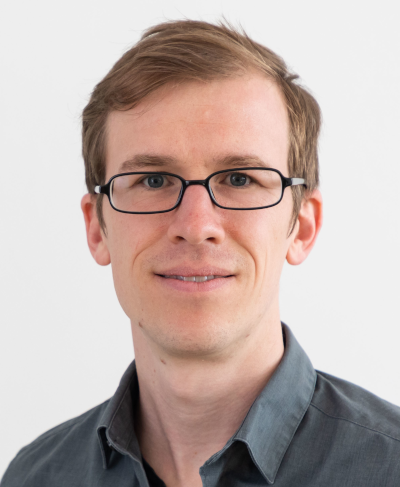}}]{Bernhard
C. Geiger}
(S'07, M'14, SM'19) received the Dipl.-Ing. degree in electrical engineering (with distinction) in 2009 and the Dr. techn. degree in electrical and information engineering (with distinction) from Graz University of Technology, Austria, in 2014.
In 2009 he joined the Signal Processing and Speech Communication Laboratory at Graz University of Technology as a Project Assistant and became a Research and Teaching Associate at the same lab in 2010. 
He was a Senior Scientist and Erwin Schr\"odinger Fellow at the Institute for Communications Engineering, Technical University of Munich, Germany from 2014 to 2017 and a postdoctoral researcher at the Signal Processing and Speech Communication Laboratory, Graz University of Technology, Austria from 2017 to 2018. 
He is currently a Senior Researcher at Know-Center GmbH, Graz, Austria, where he also leads the Machine Learning Group within the Knowledge Discovery Area. 
His research interests include information theory for machine learning, theory-assisted machine learning, and information-theoretic model reduction for Markov chains and hidden Markov models.
\end{IEEEbiography}

\EOD

\end{document}